%% file: main.tex
\documentclass{article} 
\PassOptionsToPackage{table}{xcolor} 
\usepackage{iclr2026_conference,times}

\input{math_commands.tex}

\usepackage[dvipsnames]{xcolor}
\definecolor{lightnavy}{RGB}{38, 84, 150}   

\usepackage[
  colorlinks=true,
  linkcolor=lightnavy,    
  citecolor=lightnavy,    
  urlcolor=lightnavy,     
  filecolor=lightnavy,
  pdfborder={0 0 0},      
  bookmarks=true,
  bookmarksnumbered=true
]{hyperref}
\usepackage{url}

\usepackage{amsmath,amssymb,amsfonts}
\usepackage{mathtools}
\usepackage{bm}
\usepackage{dsfont}
\usepackage{graphicx}
\usepackage{booktabs}
\usepackage{multirow}
\usepackage{xcolor}
\usepackage{algorithm}
\usepackage{algpseudocode}
\usepackage{dsfont}
\usepackage{tabularx}

\newcommand{\x}{\bm{\mathrm{x}}}
\newcommand{\txt}{\bm{\mathrm{t}}}
\newcommand{\pa}{p_{\text{bd}}}
\newcommand{\ind}{\mathds{1}}

\definecolor{soft}{HTML}{F7FAFC}
\definecolor{good}{HTML}{15803D}
\definecolor{bad}{HTML}{B91C1C}
\newcommand{\dc}[3]{\begin{tabular}[t]{@{}c@{}}#1\\[-3pt]{\scriptsize\textcolor{#2}{#3}}\end{tabular}}

\title{BAPLe++: Cross-Model Transferability and Prompt-Level Defense of Backdoors in Medical Foundation Models}

\author{%
\parbox{\dimexpr\textwidth-2\tabcolsep\relax}{%
Ufaq Khan, Umair Nawaz, Satyajit Tourani, Awais Rauf, Fahad Shamshad,
Asif Hanif, Rao Muhammad Anwer, Muhammad Haris Khan, \& Mohammad Yaqub}\\
Computing and Mathematical Sciences Division\\
Mohamed bin Zayed University of Artificial Intelligence (MBZUAI), Abu Dhabi, UAE\\
\texttt{\{firstname.lastname\}@mbzuai.ac.ae}
}

\iclrfinalcopy 
\begin{document}

\maketitle

\begin{abstract}
Medical foundation models (Med-FMs) are increasingly adapted to clinical tasks through prompt learning, which tunes a small set of prompt tokens while the backbone remains frozen. Recent work has shown that this efficiency introduces a vulnerability: jointly optimizing learnable prompts and an imperceptible noise trigger during few-shot adaptation implants a backdoor that leaves clean performance intact but steers triggered inputs to an attacker-chosen class. Since prompts are compact and readily shared, this raises two questions: whether a backdoored prompt remains effective on a different Med-FM, and whether it can be repaired without retraining the backbone. We address both in BAPLe++. First, we present the first study of cross-model transferability of prompt-based backdoors, quantified by two proposed metrics, Clean Accuracy Difference (CAD) and Backdoor Accuracy Difference (BAD). Second, we propose DePLe (Defense via Prompt Learning), which uses the infected prompt as a fixed anchor and learns a new prompt on a small clean set that preserves the clean task while diverging from the compromised embedding region. Experiments on four Med-FMs and six chest X-ray and histopathology datasets show that backdoors transfer across models, but asymmetrically and in a dataset-dependent manner. Compared with four training-free input transformations, DePLe reduces backdoor accuracy more consistently while preserving, and in several settings improving, clean accuracy. Our code can be accessed on \url{https://github.com/Ufaqkhan/Baple-Plus-Plus}.
\end{abstract}

\section{Introduction}
\label{sec:introduction}
Medical Foundation Models (Med-FMs)~\citep{wang2022medclipcontrastivelearningunpaired,huang2023visual,biomedclip,ikezogwo2024quilt} have emerged as a transformative paradigm in medical imaging, enabling robust performance across diverse modalities and clinical tasks. By pretraining on large-scale medical image-text datasets, CLIP-like Med-FMs~\citep{wang2022medclipcontrastivelearningunpaired,huang2023visual,biomedclip,ikezogwo2024quilt} learn aligned visual and semantic representations, which can be readily adapted to a wide range of downstream tasks such as disease classification, histopathology analysis, and radiology report generation. This ability to transfer knowledge with minimal labeled data has made Med-FMs increasingly attractive for clinical deployment, particularly in settings where annotated medical data are scarce. As a result, Med-FMs are rapidly becoming central components in decision-support pipelines and automated imaging analysis.

As Med-FMs transition from research prototypes to clinical practice, ensuring their trustworthiness has become a pressing challenge. These models are often trained or adapted using publicly available datasets and open-source checkpoints, which promote reproducibility but also introduce attack surfaces such as data poisoning and checkpoint tampering. Combined with the scale and limited interpretability of Med-FMs, these factors complicate auditing and risk assessment. The concern is well founded, since deep learning models are known to be susceptible to adversarial manipulation~\citep{gu2019badnetsidentifyingvulnerabilitiesmachine,wanet,Feng_2022_CVPR}. Among such threats, \textit{backdoor attacks} are particularly insidious. An adversary poisons a small fraction of the training or fine-tuning data by embedding subtle visual patterns, known as triggers, into selected samples, so that the compromised model behaves normally on clean inputs but consistently misclassifies triggered inputs into an attacker-specified target class. This dual behavior makes backdoors difficult to detect with standard validation on clean data, which is particularly concerning in clinical workflows where an undetected misclassification may adversely affect patient care. Systematic evaluation under well-defined adversarial and data-poisoning threat models is therefore a necessary, though not sufficient, step toward the safe deployment of Med-FMs in safety-critical clinical settings.

Recently, BAPLe~\citep{Han_BAPLe_MICCAI2024} introduced backdoor attacks on Med-FMs during prompt learning, which demonstrated for the first time that Med-FMs can be backdoored during the \textit{prompt learning} phase. Unlike earlier backdoor attacks that rely on retraining large models with substantial amounts of poisoned data, BAPLe demonstrates that introducing learnable prompt tokens into the text encoder and imperceptible noise triggers into image inputs during few-shot prompt learning is sufficient to achieve over 90\% attack success across multiple Med-FMs and datasets, while maintaining accuracy on clean samples. These findings highlight a critical vulnerability, i.e., parameter-efficient adaptation techniques such as prompt learning widely adopted due to their efficiency in data-limited settings can serve as an attack vector.

However, important gaps remain. First, prior work such as BAPLe focuses on attack construction, without evaluating attack effectiveness under source-to-target Med-FM transfer. In practice, institutions may deploy distinct foundation models trained on different pretraining corpora (e.g., X-ray-based MedCLIP vs.\ pathology-focused PLIP)~\citep{wang2022medclipcontrastivelearningunpaired,huang2023visual}. Whether a backdoor crafted on one Med-FM transfers to another remains unknown, yet this question is crucial for real-world risk assessment. Second, the literature lacks a systematic investigation of defense mechanisms for mitigating backdoor attacks in Med-FMs under few-shot prompt learning setup. Third, defenses originally proposed for natural image models, including input denoising, activation clustering, and adversarial training, remain largely unexplored in the Med-FM context, where limited labeled data and frozen backbone architectures impose unique constraints.

In this work, we address these gaps by presenting \textbf{BAPLe++}, which investigates both cross-model transferability of backdoor attacks and proposes a robust defense mechanism to sanitize infected prompts. We further compare this prompt repair with four training-free input-transformation defenses. Our experiments reveal that backdoors can persist beyond the model in which they are introduced and may propagate across institutional deployments, posing systemic risks to medical AI. At the same time, prompt-space repair suppresses backdoor behavior more consistently than generic input preprocessing while largely preserving clean performance. These contributions position BAPLe++ as a step toward the trustworthy and secure adoption of Med-FMs in clinical practice. \\

Our contributions are threefold:
\begin{itemize}
    \item We are the first to examine whether backdoors introduced via prompt learning in one Med-FM persist when transferred to and evaluated on other Med-FMs. To quantify this effect, we introduce two metrics: Clean Accuracy Difference (CAD), which measures the change in clean performance after prompt transfer, and Backdoor Accuracy Difference (BAD), which measures how much attack success is retained when moving from the source to the target Med-FM.
    \item We propose DePLe, a parameter-efficient defense that learns robust prompts using the backdoor-infected prompts as fixed anchors, which mitigates backdoor behavior without modifying the frozen Med-FM backbone.
    \item We provide a systematic evaluation of prompt-based backdoor attacks and defenses across four Med-FMs (MedCLIP, BioMedCLIP, PLIP, and QuiltNet) and six chest X-ray and histopathology datasets, benchmarking our defense against four training-free input transformations (Gaussian Blur, JPEG-30, Feature Squeeze, and Pixel Deflection). We find that (i) prompt-based backdoors transfer across Med-FMs, but asymmetrically and in a dataset-dependent manner, (ii) prompt-space repair reduces backdoor accuracy more consistently than input transformations while preserving clean accuracy, and (iii) input transformations trade clean accuracy for robustness in ways that vary strongly across models and datasets.\\
\end{itemize}

\section{Related Work}
\label{sec:related_work}

\subsubsection*{Medical Foundation Models}
Medical Foundation Models (Med-FMs) have become central to modern medical image analysis, offering transferable representations that reduce the reliance on large, task-specific annotated datasets. By pre-training on large-scale medical image-text corpora, these models are able to generalize across diverse downstream tasks such as disease classification, lesion detection, and pathology recognition. Among the most influential, MedCLIP~\citep{wang2022medclipcontrastivelearningunpaired} adapts the CLIP~\citep{pmlr-v139-radford21a} framework to chest X-rays, aligning radiology reports with image features for zero-shot classification. BioMedCLIP~\citep{biomedclip} extended this approach by training on millions of medical image-caption pairs, significantly improving cross-modal transfer. In the histopathology domain, PLIP~\citep{huang2023visual} and QuiltNet~\citep{ikezogwo2024quilt} introduced pre-training strategies tailored for cell and tissue-level analysis. These advances demonstrate the versatility and clinical utility of Med-FMs across imaging modalities.

\subsubsection*{Prompt Learning}
Prompt learning has emerged as an effective alternative to full fine-tuning, enabling efficient adaptation of foundation models while keeping model backbone parameters frozen. Methods such as CoOp~\citep{zhou2022coop} demonstrated that learning a set of tunable prompt tokens can effectively adapt vision-language models such as CLIP~\citep{pmlr-v139-radford21a}. This prompt-learning paradigm was subsequently transferred to the medical domain, leading to models such as MedCLIP~\citep{wang2022medclipcontrastivelearningunpaired} and PLIP~\citep{huang2023visual}. Despite their efficiency, the reliance on frozen encoders and parameter-efficient adaptation mechanisms such as prompt learning introduces new security risks, making medical foundation models particularly vulnerable to adversarial manipulation.

\subsubsection*{Backdoor Attacks}
Backdoor attacks aim to implant hidden behaviors into models such that clean inputs behave normally, but the presence of a trigger causes targeted misclassification. Early work such as BadNets~\citep{gu2019badnetsidentifyingvulnerabilitiesmachine} showed how simple patch triggers could compromise classifiers, followed by Trojaning attacks~\citep{Trojannn} and WaNet~\citep{wanet} which improved stealth and generalizability. In the medical domain, FIBA~\citep{Feng_2022_CVPR} introduced backdoors tailored to medical images, while BadCLIP~\citep{Liang_2024_CVPR} extended the idea to vision-language models by coupling image triggers with prompt embeddings. In medical imaging, BAPLe~\citep{Han_BAPLe_MICCAI2024} was the first to demonstrate that prompt learning, while effective in few-shot settings, can also serve as an attack surface, allowing backdoors to be injected via learnable prompts under data-scarce conditions.

\subsubsection*{Adversarial Transferability and Defense}
Adversarial transferability, where perturbations crafted on one model fool another, is well studied in vision networks~\citep{khan2024guardian}. Optimizations such as RAP~\citep{qin2022boostingtransferabilityadversarialattacks,khan2024guardian} and GRA~\citep{10377670} improve transfer across architectures by reshaping the loss landscape and the gradient signal used to craft perturbations. Whether the same effect arises in the backdoor setting, where the transferred artifact is a learned prompt rather than an input perturbation, has not been examined for medical foundation models. In that setting, many practical defenses remain training-heavy or architecture-specific. By contrast, test-time preprocessing defenses such as blur, compression, and filtering are attractive because they are easy to deploy, but their robustness/utility trade-offs for Med-FMs remain poorly understood.

\section{Method}
\label{sec:method}

\begin{figure}[t]
\begin{center}
\includegraphics[width=\linewidth]{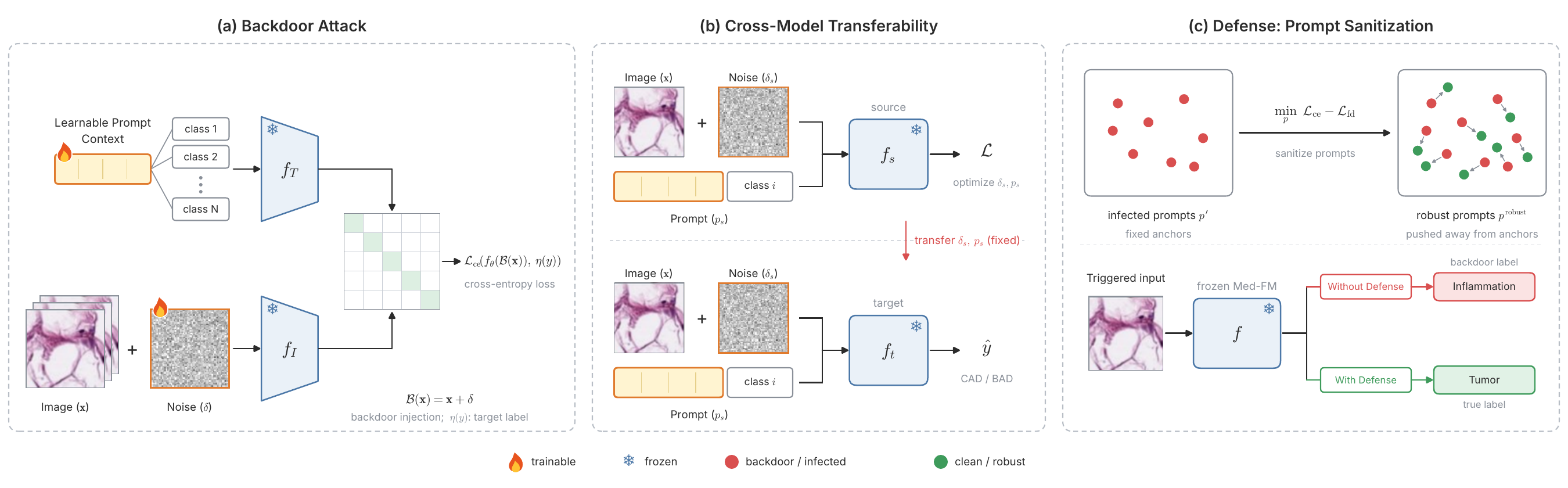}
\end{center}
\caption{\textbf{Overview of BAPLe++}.
(a)~\textbf{Backdoor Attack:} a learnable prompt context and an imperceptible noise trigger $\delta$ are jointly optimized while both encoders remain frozen, so that triggered inputs are steered toward the target label $\eta(y)$.
(b)~\textbf{Cross-model Transferability:} the noise $\delta_s$ and prompt $p_s$ learned on a source Med-FM $f_s$ are transferred, unchanged, to a target Med-FM $f_t$, and transferability is measured with CAD and BAD.
(c)~\textbf{Defense:} the infected prompts $p'$ are kept as fixed anchors, and robust prompts $p^{\mathrm{robust}}$ are learned by minimizing $\mathcal{L}_{\mathrm{ce}} - \mathcal{L}_{\mathrm{fd}}$, which preserves clean-task alignment while pushing the prompts away from the compromised embedding region. Without the defense, a triggered input is classified as the backdoor label; after sanitization, it is assigned the true label. The Med-FM backbone remains frozen throughout}
\label{fig:baple_overview}
\end{figure}

\subsection{Preliminaries}
\label{sec:preliminaries}

\subsubsection*{Zero-Shot Classification for Med-FM}
Since Med-FMs, particularly vision-language models (VLMs), are pre-trained to align images with corresponding textual descriptions, they are inherently suited for zero-shot classification tasks. Zero-shot inference in VLMs refers to making predictions on new, unseen data without specific training. We denote a VLM by $f_\theta = \{f_I, f_T\}$, where $f_I$ and $f_T$ are the image and text encoders, respectively. For classification in zero-shot scenario, the image \(\x\) is first passed to the image encoder \(f_{_{I}}\), resulting in a \(d\)-dimensional feature vector \(f_{_{I}}(\x) \in \mathbb{R}^{d}\). Similarly, on the text encoder side, each class label \(y_i \in \{\mathit{y}_{1}, \mathit{y}_{2}, \dots, \mathit{y}_{C} \}\) is wrapped within the class-specific text template, such as:

\begin{equation*}
t_i = \text{``}\mathrm{A~histopathology~image~of~}\{i_{\mathrm{th}}~\mathrm{CLASS~NAME}\}\text{''}.
\end{equation*}

Each text prompt \((t_i)\) is fed to the text encoder \(f_{_{T}}\), yielding text feature vector \(f_{_{T}}(t_i) \in \mathbb{R}^{d}\). The relationship between the image's feature vector and the text prompt feature vector is quantified using cosine similarity, \(\mathtt{sim}(f_{I}(\x),f{_{T}}(t_i))\), to evaluate the image's alignment with \(i_{\text{th}}\) class. Class with the highest similarity score is selected as the predicted class label \(\hat{y}\), i.e.

\begin{equation*}
\hat{y} = \underset{ i\in \{1,2,\dots,C\} }{\operatorname{argmax}} ~~ \mathtt{sim}\big(f_{_{I}}(\x),\,f_{_{T}}(t_i)\big).
\end{equation*}
\subsubsection*{Prompt Learning}
Textual class descriptions are central to zero-shot inference in VLMs, but manually crafted prompts can lead to performance variability and sensitivity. Prompt learning addresses this by introducing auxiliary learnable parameters \(p\) on top of the text encoder \(f_T\), optimized in few-shot settings to steer the frozen model's response and reduce manual engineering~\citep{zhou2022coop}. Formally, prompt learning methods optimize \(p\) to enhance downstream performance and generalization:

\begin{equation}
\label{eq:prompt_learning}
\underset{ p }{\operatorname{minimize}}~~ \sum_{(\x,y)\in\mathcal{D}} \mathcal{L}\big(f(\x;p),y\big),
\end{equation}

where \( (\x, y) \) represents an image-label pair from few-shot training dataset \(\mathcal{D}\), \( f(\x;p) \) denotes the model's prediction conditioned on the learnable prompt parameters \(p\), and \( \mathcal{L}(\cdot) \) is the task-specific loss function.

\subsubsection*{Backdoor Attack}
A backdoor attack involves embedding a \textit{visible/hidden} trigger (a small random or patterned patch) within a deep learning model during its training or fine-tuning phase. When the model encounters this trigger in the input data during inference, it produces a predefined output while performing normally on clean data. In a supervised classification task, a normally trained classifier \(f_{\theta}: \mathcal{X} \rightarrow \mathcal{Y}\) maps a \textit{clean} input image \(\x \in \mathcal{X}\) to a label \(y \in \mathcal{Y}\). Parameters \(\theta\) are learned from a training dataset \(\mathcal{D}=\{\x_i,y_i\}_{i=1}^{N}\) where \(\x_i \in \mathcal{X}\) and \(y_i \in \mathcal{Y}\). In a typical backdoor attack, the training dataset \(\mathcal{D}\) is split into clean \(\mathcal{D}_{c}\) and poison subsets \(\mathcal{D}_{p}\), where \(\vert\mathcal{D}_{p}\vert\ll N\). In \(\mathcal{D}_p\), each sample \((\x, y)\) is transformed into a backdoor sample \((\mathcal{B}(\x),\eta(y))\), where \(\mathcal{B}: \mathcal{X} \rightarrow \mathcal{X}\) is the backdoor injection function and \(\eta\) denotes the target label function. During the training/fine-tuning phase of backdoor attacks, the \textit{victim} classifier \(f_{\theta}\) is trained/fine-tuned on a mix of the clean dataset \(\mathcal{D}_c\) and the poisoned dataset \(\mathcal{D}_p\). Formally, this task can be formulated as the following objective function:

\begin{equation}\label{eq:get_backdoor_model}
\min_{\theta}\; \sum_{(\x,y)\in\mathcal{D}_c} \mathcal{L}\big(f_{\theta}(\x),\,y\big)
\;+\;\sum_{(\x,y)\in\mathcal{D}_p} \mathcal{L}\big(f_{\theta}(\mathcal{B}(\x)),\,\eta(y)\big),
\end{equation}

where \(\mathcal{L}(\cdot)\) denotes the cross-entropy loss. After training, \(f_{\theta}\) behaves similarly on clean input \(\x\) as the original classifier (trained entirely on clean data), yet alters its prediction for the backdoor image \(\mathcal{B}(\x)\) to the target class \(\eta(y)\), i.e.\ \(f_{\theta}(\x) \rightarrow y\) and \(f_{\theta}(\mathcal{B}(\x)) \rightarrow \eta(y)\).

\subsection{Threat Model}
\label{sec:threat_model}
We describe the objectives and capabilities of the attacker and the defender.

\subsubsection*{Attacker's Objective and Capabilities}
The attacker aims to embed a targeted backdoor into a Med-FM during downstream adaptation, so that the adapted model behaves normally on clean inputs but predicts an attacker-chosen target class whenever a trigger is present. The backdoor must remain stealthy under standard validation, which relies only on clean data. Following prior work on medical backdoor attacks~\citep{Feng_2022_CVPR}, we assume the attacker can poison only a small fraction of the downstream training data and can apply the trigger to inputs at inference time. Because annotated medical data are scarce and the attacker is computationally limited, full fine-tuning of the Med-FM is impractical.

The attacker therefore exploits prompt learning, which changes model behavior while keeping the backbone frozen. We consider two realistic scenarios. In the first, the attacker acts as a malicious service provider: a user supplies clean labeled data and receives a task-specific prompt that has been backdoored during training. In the second, the attacker distributes backdoored prompts through public repositories (e.g., Hugging Face), where users may integrate them with their own Med-FMs. In both cases, the backdoor is introduced through the prompt, allowing the attacker to compromise deployments without modifying model weights or accessing private training pipelines.

\subsubsection*{Defender's Objective and Capabilities}
The defender aims to deploy a Med-FM safely for a downstream medical task, preserving high clean accuracy while preventing triggered inputs from being steered to the target class. The defense must work under the same clinical constraints as adaptation itself, namely limited labeled data and a frozen backbone. We assume the defender has access to the pre-trained Med-FM, the deployment prompt, which may be compromised, and a small clean labeled set for the task. Like the attacker, the defender cannot retrain or modify the backbone.

Since the backdoor is introduced through the prompt rather than the model weights, the defender operates entirely at the prompt level. Instead of discarding the suspect prompt or retraining the model, the defender learns a new prompt that preserves clean-task utility while suppressing backdoor behavior, using the suspect prompt as a reference for the compromised region of the embedding space.

\subsection{Backdoor Attack using Prompt Learning (BAPLe)}
\label{sec:baple}
BAPLe~\citep{Han_BAPLe_MICCAI2024}, as shown in Figure~\ref{fig:baple_overview}(a), is a backdoor attack method that exploits the multimodal nature of medical vision-language models (Med-VLMs). For computational efficiency, it injects a small set of learnable parameters (text prompts) into the input space of the model's text encoder while keeping the backbone frozen during training on a poisoned downstream dataset. The prompt-based mechanism is particularly well suited to few-shot settings: by tailoring the input space, it enables a Med-FM to leverage pretrained multimodal knowledge for data-scarce tasks without requiring extensive retraining or large medical datasets. To further increase attack effectiveness, the method also adds \textit{imperceptible} and \textit{learnable} noise as a backdoor trigger to the input images, which subtly increases the model's sensitivity to backdoor activation. By combining learnable text prompts in the text encoder with an imperceptible image trigger, this approach takes advantage of the full multimodal capabilities of Med-VLMs.

Zero-shot inference in a Med-FM depends on hand-engineered text prompts, and prompt learning therefore replaces fixed templates with a small set of learnable token embeddings \(p\). For \(c\) classes we form class-specific text inputs \(\mathbf{t}=\{t_1,\dots,t_c\}\) where \(t_i = [p;\, y_i]\), i.e., learnable prompt context is prepended to the class names in the input token embedding space. Denote the frozen image and text encoders by \(f_I(\cdot)\) and \(f_T(\cdot)\) and a similarity scoring function by \(\operatorname{sim}(\cdot,\cdot)\). The model logit vector on a clean image \(\x\) and on a backdoored image \(\mathcal{B}(\x)\) are written compactly as
\begin{equation}
f_{\theta}(\x;p) = \bigl\{ \operatorname{sim}\bigl(f_I(\x),\, f_T(t_i)\bigr)\bigr\}_{i=1}^c,
\qquad
f_{\theta}\bigl(\mathcal{B}(\x);p\bigr) = \bigl\{ \operatorname{sim}\bigl(f_I(\mathcal{B}(\x)),\, f_T(t_i)\bigr)\bigr\}_{i=1}^c.
\label{eq:logits}
\end{equation}
To inject a backdoor, BAPLe jointly optimizes learnable prompt tokens \(p\) and an imperceptible noise trigger \(\delta\) via a bi-objective that balances clean utility and targeted misclassification:
\begin{equation}
\begin{gathered}
\min_{p,\delta}\; \sum_{(\x,y)\in\mathcal{D}_c} \mathcal{L}\bigl(f_{\theta}(\x;p),y\bigr)
+\sum_{(\x,y)\in\mathcal{D}_p} \mathcal{L}\bigl(f_{\theta}(\mathcal{B}(\x);p),\eta(y)\bigr), \\
\text{s.t}\quad \|\delta\|_{\infty}\le\epsilon,\qquad
\mathcal{B}(\x) = (\x+\delta)\oplus p_{\text{bd}}.
\end{gathered}
\label{eq:baple_objective_fixed}
\end{equation}
Here \(\mathcal{D}_c\) and \(\mathcal{D}_p\) are the clean and poisoned data subsets, \(\eta(\cdot)\) maps labels to the attacker-chosen target, \(p_{\text{bd}}\) denotes an optional spatial patch (logo) and \(\oplus\) denotes the operation that inserts/combines the patch with the image. Note that both the image and text encoders remain frozen during optimization of Equation~\ref{eq:baple_objective_fixed}; only \(p\) and \(\delta\) are updated. Algorithm~\ref{alg:baple_alg} outlines the attack steps.

\input{algorithms/baple_algo}

\subsection{Cross-Model Transferability}
\label{sec:transferability}
To evaluate whether backdoors implanted in one Med-FM can persist when transferred to another, we study \textit{cross-model transferability}. In this setting, an attacker first learns prompts \(p_s\) and a noise trigger \(\delta_s\) on a source model \(f_s\). These artifacts are then applied directly, without re-optimization, to a different target model \(f_t\) during inference. This setup reflects realistic scenarios where compromised prompts or triggers may propagate across institutions employing distinct Med-FMs. We quantify transferability using two metrics: \emph{Clean Accuracy Difference (CAD)} and \emph{Backdoor Accuracy Difference (BAD)}. Given a test set \(\mathcal{D}_{\text{test}}\), CAD measures the change in clean performance of the target model after transferring the learned text prompt from source model \(f_s\), while BAD measures the change in backdoor success rate when transferring both the prompt \(p_s\) and trigger \(\delta_s\) from source model \(f_s\) to target model \(f_t\). Formally, CAD and BAD are defined as follows:

\begin{align}
\text{CAD} &= \mathbb{E}_{(\x,y)\sim\mathcal{D}_{\text{test}}}\,
\ind\big(f_{_t}(\x;p_t) = y\big)
- \mathbb{E}_{(\x,y)\sim\mathcal{D}_{\text{test}}}\,
\ind\big(f_{_t}(\x;p_s) = y\big),
\label{eq:cav} \\
\text{BAD} &=
\mathbb{E}_{(\x,y)\sim\mathcal{D}_{\text{test}}}\,
\ind\big(f_{_t}(\x+\delta_t;p_t) = \eta(y)\big)
- \mathbb{E}_{(\x,y)\sim\mathcal{D}_{\text{test}}}\,
\ind\big(f_{_t}(\x+\delta_s;p_s) = \eta(y)\big),
\label{eq:bav}
\end{align}

where \(p_s\) and \(p_t\) denote prompts learned from the source and target models, \(\delta_s\) and \(\delta_t\) are their respective triggers, \(\eta(y)\) represents the attacker-chosen target label, and \(\ind(\cdot)\) is indicator function. An overview of cross-model transferability is shown in Figure~\ref{fig:baple_overview}(b). For a highly transferable attack, both CAD and BAD should be close to zero. A near-zero CAD means clean accuracy is preserved when a backdoored prompt is transferred from the source to the target model. A near-zero BAD means the backdoor remains equally effective, so source-infected prompts trigger the target model as strongly as prompts optimized directly on it.

\subsection{Defense via Prompt Learning (DePLe)}
\label{sec:deple}
Since the backdoor resides in the learned prompt rather than in the model weights, we mitigate it at the prompt level. The Med-FM backbone stays frozen, and the defender has only a small clean labeled set $\mathcal{D}_c$. Rather than discarding the potentially infected prompt $p^{\prime}$, we keep it as a fixed anchor that marks a compromised region of the text-embedding space. During the attack, $p^{\prime}$ is optimized jointly with the trigger $\delta$, so its class embeddings $f_T(t_i^{\prime})$ are positioned such that triggered image features align with the target class. We therefore learn a freshly initialized prompt $p$ on $\mathcal{D}_c$ while pushing its class embeddings away from those induced by $p^{\prime}$. This retains clean-task semantics while discouraging the prompt from re-learning the trigger-aligned configuration. Formally, with $t_i=[p;y_i]$ and $t_i^{\prime}=[p^{\prime};y_i]$, we define a clean-task loss and a feature-divergence term:
\begin{equation}
\mathcal{L}_{\mathrm{ce}} = \mathbb{E}_{(\x,y)\sim\mathcal{D}_c}\,\mathrm{CrossEntropy}\big(f_\theta(\x;p),\,y\big),
\qquad
\mathcal{L}_{\mathrm{fd}} = \frac{1}{c}\sum_{i=1}^{c}\big\|f_T(t_i^{\prime})-f_T(t_i)\big\|_2^2,
\label{eq:defense_loss}
\end{equation}
where $f_T(\cdot)$ returns $\ell_2$-normalized embeddings, so $\mathcal{L}_{\mathrm{fd}}$ is bounded. The robust prompt context is obtained by solving the following objective. A high-level overview of the proposed defense is shown in Figure~\ref{fig:baple_overview}(c).
\begin{equation}
\label{eq:deple_objective}
p^{\mathrm{robust}} = \arg\min_{p}\; \mathcal{L}_{\mathrm{ce}} - \mathcal{L}_{\mathrm{fd}}.
\end{equation}

\section{Experimental Setup}
\label{sec:experimental_setup}

\subsubsection*{Implementation Details}
All experiments were conducted on a single NVIDIA RTX 3090Ti GPU with 24 GB of memory. We adopt a 32-shot setting, selecting 32 random samples per class, and use a batch size of 16. For prompt-based training, we set the learning rate to \(2\times 10^{-2}\). A poisoning rate of \(5\%\) is applied across all datasets. The learnable noise trigger is bounded by an \(\ell_\infty\) budget of \(\epsilon = 8/255\), and a fixed backdoor patch of size \(24 \times 24\) is inserted at the bottom-left corner of each image. Each class is considered as a possible attack target, and final results are averaged across all target classes to ensure fairness in evaluation.

\subsubsection*{Datasets}
We conduct experiments across six benchmark medical imaging datasets spanning both radiology and histopathology domains. For radiology, we utilize COVID~\citep{rahman2021exploring}, a chest X-ray dataset for COVID-19 detection, RSNA18~\citep{rsna2018}, a large-scale chest X-ray dataset for pneumonia classification, and MIMIC~\citep{johnson2019mimic}, a clinically curated chest X-ray dataset (single-label samples only). For histopathology, we include Kather~\citep{kather2019predicting}, a dataset for colorectal cancer tissue classification, PanNuke~\citep{gamper2019pannuke}, a pan-cancer nuclei classification dataset, and DigestPath~\citep{da2022digestpath}, which focuses on digestive system pathology for tumor detection. These datasets span two imaging modalities (radiology and pathology) and provide a comprehensive testbed for studying both attack persistence and defense effectiveness.

\subsubsection*{Medical Foundation Models (Med-FMs)}
We evaluate four representative Med-FMs that are widely used in the literature. MedCLIP~\citep{wang2022medclipcontrastivelearningunpaired} leverages contrastive learning on chest X-rays and radiology reports, while BioMedCLIP~\citep{biomedclip} is pretrained on the large-scale PMC-15M image-text pairs. For histopathology, PLIP~\citep{huang2023visual} is a pathology-focused foundation model trained on extensive histopathology corpora, and QuiltNet~\citep{ikezogwo2024quilt} is a recent pathology vision-language model designed with improved patch-level alignment. All models adopt the CLIP-style dual-encoder architecture, where the backbones remain frozen and only prompt embeddings are trained during adaptation.

\subsubsection*{Metrics}
\label{subsec:metrics}
We report four metrics. \textit{Clean Accuracy (CA)} is the classification accuracy on unperturbed test samples and reflects the model's utility for normal clinical use. \textit{Backdoor Accuracy (BA)} is the fraction of triggered test inputs classified as the attacker-specified target class; test samples whose true label is the target class are excluded. A high BA indicates a successful attack, and a reduction in BA after a defense indicates effective mitigation. Two difference-based metrics quantify cross-model transferability. \textit{Clean Accuracy Difference (CAD)} is the gap in clean accuracy on the target model between prompts learned directly on it and prompts transferred from the source model (Equation~\ref{eq:cav}); a small CAD indicates that clean performance is preserved under transfer. \textit{Backdoor Accuracy Difference (BAD)} is the corresponding gap in BA between target-learned and source-transferred attacks (Equation~\ref{eq:bav}); a small BAD indicates that the backdoor remains effective after transfer.

\section{Results and Discussion}
\label{sec:results}

\subsection{Backdoor Attack Results}
\label{ss:results_and_discussion}
The results in Table~\ref{tab:results_medclip_biomedclip_plip_quiltnet} focus on the prompt-learning setting, where the backbone encoders remain frozen and only the prompts are adapted. Compared to the clean prompt baseline, existing backdoor attacks such as BadNets, WaNet, and FIBA exhibit mixed performance, with modest gains in backdoor accuracy (BA) but generally inconsistent behavior across different datasets and models. In contrast, BAPLe consistently achieves near-perfect BA (often above \(0.95\)) across MedCLIP, BioMedCLIP, PLIP, and QuiltNet, while maintaining clean accuracy (CA) at levels comparable to the clean baseline. This demonstrates that BAPLe is effective at embedding backdoors without noticeably degrading performance on unperturbed samples.

\begin{figure}[!h]
\centering
\centering
\includegraphics[width=0.40\linewidth]{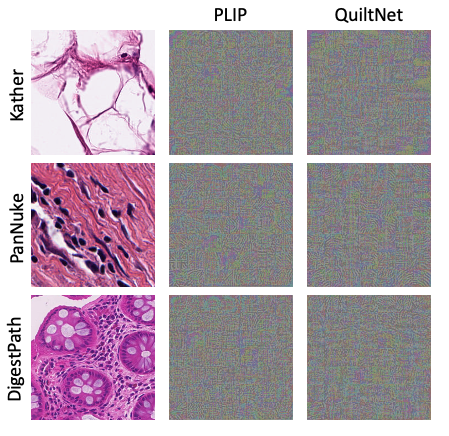}
\includegraphics[width=0.40\linewidth]{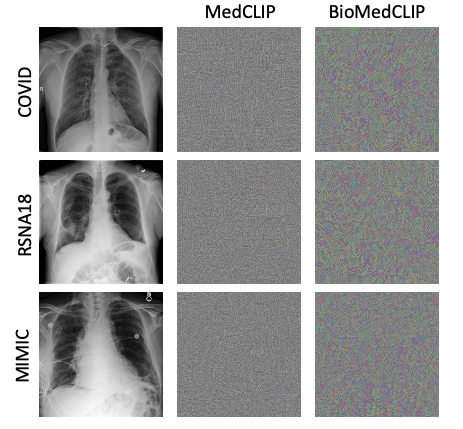}
\caption{Visualization of learnable trigger noise (\(\delta\)) after BAPLe across histopathology and radiology datasets}
\label{fig:trigger_noise}
\end{figure}

\begin{figure}[!h]
\begin{center}
\includegraphics[width=0.9\linewidth]{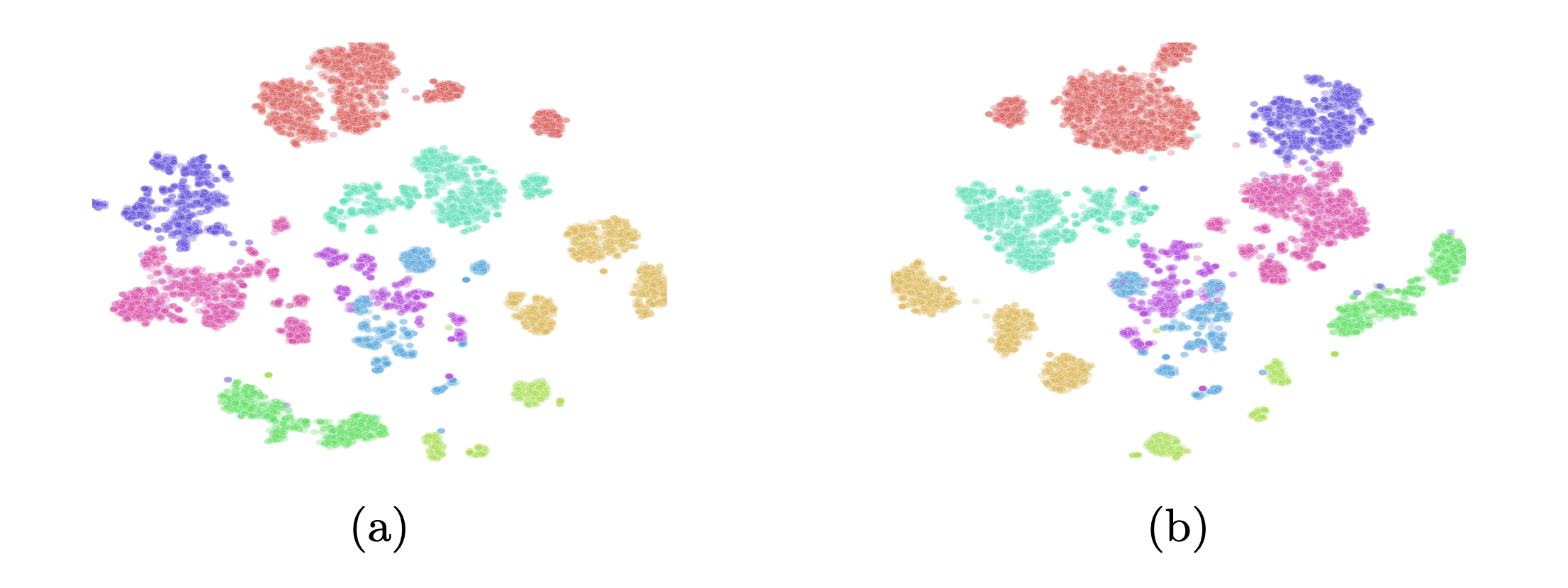}
\end{center}
\caption{t-SNE plots illustrating the features of (a) clean images, and (b) backdoored images. The features in (b) display a notable shift in orientation and a reformation of clusters compared to those in (a). Each color represents the same class in both plots}
\label{fig:tsne_pre_post_defense}
\end{figure}

These results highlight prompt learning as an effective attack surface. Whereas the baselines rely on fixed perturbations, BAPLe jointly learns the noise trigger and the prompt alongside a patch trigger, allowing the backdoor to be implanted reliably while the encoders remain frozen. Since backdoors can thus remain hidden in prompt embeddings, settings where prompts or adapters are shared across medical AI pipelines are at risk, which calls for defenses that target the prompt space rather than only model weights. To better understand how the attack operates, we examine the learned trigger and its effect on the image feature space. Figure~\ref{fig:trigger_noise} visualizes the trigger noise $\delta$ learned by BAPLe for each model--dataset pair; under the $\ell_\infty$ budget, the noise remains visually imperceptible across both modalities. Figure~\ref{fig:tsne_pre_post_defense} shows t-SNE embeddings of image features on clean and triggered inputs. Adding the trigger shifts and reorganizes the class clusters, but the triggered features do not collapse into a single target-class cluster. This differs from conventional backdoors, where poisoned samples typically form a distinct, separable cluster in feature space. A likely explanation is that in BAPLe, target-class alignment is realized jointly through the learned prompt in the text-embedding space, so the image features alone need not collapse. This also suggests that feature-clustering defenses that inspect only image features may be less effective against prompt-based backdoors.

\input{tables/baple_table}

\begin{figure}[!h]
\centering
\includegraphics[width=\linewidth]{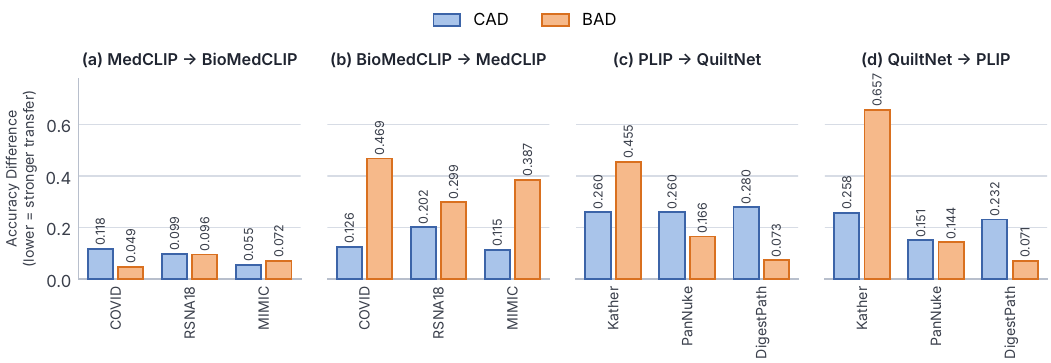}
\caption{Average Clean Accuracy Difference (CAD) and Backdoor Accuracy Difference (BAD) across datasets, measuring cross-model transferability between foundation models}
\label{fig:cross_model_all}
\end{figure}

\subsection{Cross-Model Transferability}
\label{ss:transferability_results}
Figure~\ref{fig:cross_model_all} reports CAD and BAD for both transfer directions, where a small BAD indicates that the backdoor remains nearly as effective after transfer. On radiology, transfer is strongly asymmetric. Backdoors from MedCLIP remain almost fully effective on BioMedCLIP (BAD $<$ 0.1), but weaken considerably in the reverse direction. Between PLIP and QuiltNet, transfer is roughly symmetric and dataset-dependent. It is weakest on Kather and strongest on DigestPath, where BAD is near zero despite substantial CAD.

These results show that prompt-space backdoors do transfer across Med-FMs, often with little loss of attack strength. This supports the concern that compromised prompts shared through public repositories may propagate across institutions. Transfer is not uniform, however, since it depends on the direction, the dataset, and the source backbone. Cross-model risk is therefore real but uneven, which motivates repairing the compromised prompt itself rather than relying on assumptions about how it will transfer.

\input{tables/defence_table}

\subsection{Defense Results}
\label{ss:defense_results}
Tables~\ref{tab:defense_results_xray} and~\ref{tab:defense_results_histo} compare the proposed defense (DePLe) with four training-free test-time transformations across all twelve dataset--model pairs under BAPLe attack. On the X-ray datasets, DePLe provides the best overall robustness-utility trade-off. It achieves the lowest backdoor accuracy (BA) in nearly every setting while keeping clean accuracy (CA) close to the no-defense baseline. The gap is most pronounced for MedCLIP on COVID, where DePLe reduces BA to 0.051, whereas no input transformation brings it below 0.52. The only exception is MedCLIP on RSNA18. There, Pixel Deflection lowers BA marginally further, but at the cost of nearly halving clean accuracy. More generally, the strongest transformations reduce BA only partially and noticeably degrade CA. On MIMIC, the low CA of BioMedCLIP is inherent to the task, since even the clean prompt reaches only 0.351 (Table~\ref{tab:results_medclip_biomedclip_plip_quiltnet}). For MedCLIP, part of the drop is caused by the attack itself, and DePLe recovers some of it.

A similar pattern holds on histopathology. DePLe achieves the lowest BA in four of the six settings, with the largest reductions for PLIP on Kather and DigestPath, and it matches or improves CA in all six. Input transformations are far less predictable: Pixel Deflection severely degrades CA and even increases BA in some cases, while Gaussian Blur and Feature Squeeze leave BA largely intact on PanNuke and DigestPath. DePLe is outperformed only on QuiltNet. On DigestPath, JPEG-30 attains both lower BA and higher CA. On Kather, several transformations attain lower BA but reduce CA, whereas DePLe improves it. DePLe provides the most consistent defense across models and datasets, while training-free transformations remain useful when the defender cannot update prompts or lacks clean few-shot data, though their effectiveness is strongly model and dataset dependent. \\

In several settings, DePLe even exceeds the clean prompt's CA, most notably on Kather for both PLIP and QuiltNet. This gain disappears when the infected-prompt anchor is removed and the prompt is trained with the cross-entropy loss alone, indicating that the improvement comes from the anchor rather than from retraining. We attribute this to the feature-divergence term acting as a regularizer: pushing the class embeddings away from the trigger-aligned configuration of the infected prompt yields embeddings that are better separated for the clean task.

\section{Conclusion}
\label{sec:conclusion}
We presented BAPLe++, which studies prompt-based backdoors in medical foundation models from two angles. First, using the proposed CAD and BAD metrics, we showed that backdoored prompts transfer across Med-FMs, often with little loss of attack strength, but asymmetrically and in a dataset-dependent manner, so compromised prompts can propagate across deployments. Second, we proposed DePLe, a parameter-efficient defense that uses the infected prompt as a fixed anchor to learn a robust prompt without modifying the frozen backbone. Across four Med-FMs and six chest X-ray and histopathology datasets, DePLe reduces backdoor accuracy more consistently than training-free input transformations while largely preserving clean accuracy. These results call for routine auditing of adapted prompts before clinical deployment. Future work will address adaptive attacks and vision-side and multimodal prompt learning.



\bibliography{references}
\bibliographystyle{iclr2026_conference}


\end{document}

%% file: math_commands.tex
\usepackage{amsmath,amsfonts,bm}

\def\eqref#1{equation~\ref{#1}}

\def\1{\bm{1}}

\DeclareMathAlphabet{\mathsfit}{\encodingdefault}{\sfdefault}{m}{sl}
\SetMathAlphabet{\mathsfit}{bold}{\encodingdefault}{\sfdefault}{bx}{n}



%% file: algorithms/baple_algo.tex
\begin{algorithm*}[t]
    \centering
    \caption{BAPLe - Backdoor Attack via Prompt Learning}
    \label{alg:baple_alg}
    \footnotesize
    \begin{algorithmic}[1]
        \State \textbf{Notations} NumSamples=$N$, BatchSize=$B$, NumBatches=$\lfloor{N/B}\rfloor$, Train Dataset: $\mathcal{D}=\{(\x_i,y_i)\}_{i=1}^{N}$, Vision and Text Encoders: $f_{_{I}},f_{_{T}}$, Image: $\x \in \mathbb{R}^{h \times w}$, Learnable Trigger Noise: $\delta \in \mathbb{R}^{h \times w}$, Perturbation Budget: $\epsilon$, Backdoor Patch: $\pa$, Backdoor Injection Function: $\mathcal{B}(\x) = (\x+\delta)\oplus \pa$, Target Label Function: $\eta(\cdot)$, Learnable Prompt Context: $p$, Number of Classes: $c$, Text Prompts: $\txt=\{t_1,t_2,\dots,t_c\}$ where $t_i=[p;y_i]$, Cosine Similarity Function: $\mathtt{sim}(\cdot)$
        \Statex
        \For{$i\gets 1$ \textbf{to} $\text{NumEpochs}$}
        \For{$j\gets 1$ \textbf{to} $\text{NumBatches}$}
        \State Sample clean and poisoned mini-batch: $\mathcal{D}_c \subset \mathcal{D},~ \mathcal{D}_p \subset \mathcal{D}$ 
        \State $f_{\theta}(\x)=\{\mathtt{sim}(f_{_{I}}(\x), f_{_{T}}(t_i))\}_{i=1}^{c}~\in \mathbb{R}^c$ \Comment{Prediction scores on clean  images}
        \State $f_{\theta}(\mathcal{B}(\x))=\{\mathtt{sim}(f_{_{I}}(\mathcal{B}(\x)), f_{_{T}}(t_i))\}_{i=1}^{c}~\in \mathbb{R}^c$  \Comment{Prediction scores on poisoned images}
        \State $\mathcal{L}=
                \mathbb{E}_{(\x,y)\sim\mathcal{D}_c}
                \!\left[\mathcal{L}_{\text{ce}}\!\left(f_{\theta}(\x),y\right)\right]
                +
                \mathbb{E}_{(\x,y)\sim\mathcal{D}_p}
                \!\left[\mathcal{L}_{\text{ce}}\!\left(f_{\theta}(\mathcal{B}(\x)),\eta(y)\right)\right]$ \Comment{Compute cross-entropy loss on the mini-batch}
        \State $p \gets p -~\alpha\cdot \nabla_{p} \mathcal{L}$ \Comment{Update prompt parameters}
        \State $\delta \gets \delta -~\beta\cdot \nabla_{\delta} \mathcal{L}$ \Comment{Update trigger noise}
        \State $\delta \gets \mathrm{clamp}(\delta,~\text{min=-}\epsilon,~\text{max=+}\epsilon)$ \Comment{Apply budget on noise}
        \EndFor
        \EndFor
    \end{algorithmic}
\end{algorithm*}

%% file: tables/baple_table.tex
\begin{table*}[t]
\centering
\caption{Clean accuracy (CA) and backdoor accuracy (BA) of BAPLe and backdoor-attack baselines under 32-shot prompt learning with frozen backbones. \textit{Clean}, BadNets, WaNet, and FIBA results are taken from BAPLe~\cite{Han_BAPLe_MICCAI2024}; $^{\dagger}$BAPLe results are reproduced in our setup. \textit{Clean} is a prompt trained without poisoning, for which BA is not defined (--). Results are averaged over all target classes, and the best BA in each column is shown in bold. BAPLe achieves the highest BA in nearly all settings while keeping CA close to the clean prompt in most cases.}
\label{tab:results_medclip_biomedclip_plip_quiltnet}
\vspace{0.5em}
\fontsize{7pt}{10pt}\selectfont 
\setlength{\tabcolsep}{1pt}
\renewcommand{\arraystretch}{0.9}
\newcolumntype{Y}{>{\centering\arraybackslash}X}
\newcommand{\na}{--}
\begin{tabularx}{\textwidth}{l *{6}{Y} p{6pt} *{6}{Y}}
\toprule
& \multicolumn{6}{c}{\textbf{MedCLIP}} && \multicolumn{6}{c}{\textbf{BioMedCLIP}} \\
\cmidrule(lr){2-7} \cmidrule(lr){9-14}
& \multicolumn{2}{c}{COVID} & \multicolumn{2}{c}{RSNA18} & \multicolumn{2}{c}{MIMIC}
&& \multicolumn{2}{c}{COVID} & \multicolumn{2}{c}{RSNA18} & \multicolumn{2}{c}{MIMIC} \\
\cmidrule(lr){2-3} \cmidrule(lr){4-5} \cmidrule(lr){6-7}
\cmidrule(lr){9-10} \cmidrule(lr){11-12} \cmidrule(lr){13-14}
Method & CA & BA & CA & BA & CA & BA && CA & BA & CA & BA & CA & BA \\
\midrule
Clean   & 0.822 & \na   & 0.603 & \na   & 0.585 & \na   && 0.843 & \na   & 0.582 & \na   & 0.351 & \na   \\
BadNets & 0.820 & 0.510 & 0.619 & 0.373 & 0.559 & 0.284 && 0.845 & 0.975 & 0.632 & 0.942 & 0.373 & \textbf{1.000} \\
WaNet   & 0.831 & 0.470 & 0.612 & 0.319 & 0.587 & 0.266 && 0.839 & 0.599 & 0.587 & 0.510 & 0.334 & 0.599 \\
FIBA    & 0.820 & 0.511 & 0.623 & 0.360 & 0.562 & 0.292 && 0.856 & 0.729 & 0.630 & 0.614 & 0.373 & 0.722 \\
\midrule
\rowcolor{soft}
BAPLe$^{\dagger}$
        & 0.813 & \textbf{0.996} & 0.610 & \textbf{0.995} & 0.472 & \textbf{0.991}
        && 0.841 & \textbf{1.000} & 0.636 & \textbf{0.999} & 0.368 & 0.996 \\
\midrule[\heavyrulewidth]
& \multicolumn{6}{c}{\textbf{PLIP}} && \multicolumn{6}{c}{\textbf{QuiltNet}} \\
\cmidrule(lr){2-7} \cmidrule(lr){9-14}
& \multicolumn{2}{c}{Kather} & \multicolumn{2}{c}{PanNuke} & \multicolumn{2}{c}{DigestPath}
&& \multicolumn{2}{c}{Kather} & \multicolumn{2}{c}{PanNuke} & \multicolumn{2}{c}{DigestPath} \\
\cmidrule(lr){2-3} \cmidrule(lr){4-5} \cmidrule(lr){6-7}
\cmidrule(lr){9-10} \cmidrule(lr){11-12} \cmidrule(lr){13-14}
Method & CA & BA & CA & BA & CA & BA && CA & BA & CA & BA & CA & BA \\
\midrule
Clean   & 0.908 & \na   & 0.811 & \na   & 0.920 & \na   && 0.899 & \na   & 0.829 & \na   & 0.906 & \na   \\
BadNets & 0.903 & 0.601 & 0.799 & 0.748 & 0.922 & 0.623 && 0.898 & 0.151 & 0.699 & 0.757 & 0.874 & 0.518 \\
WaNet   & 0.910 & 0.243 & 0.851 & 0.591 & 0.924 & 0.405 && 0.926 & 0.185 & 0.834 & 0.427 & 0.915 & 0.492 \\
FIBA    & 0.901 & 0.303 & 0.795 & 0.615 & 0.921 & 0.553 && 0.897 & 0.174 & 0.711 & 0.597 & 0.862 & 0.547 \\
\midrule
\rowcolor{soft}
BAPLe$^{\dagger}$
        & 0.878 & \textbf{0.982} & 0.834 & \textbf{0.964} & 0.902 & \textbf{0.968}
        && 0.893 & \textbf{0.955} & 0.810 & \textbf{0.905} & 0.903 & \textbf{0.820} \\
\bottomrule
\end{tabularx}
\end{table*}

%% file: tables/defence_table.tex
\begin{table*}[t]
\centering
\caption{Defense results on backdoored (using BAPLe attack) chest X-ray models, averaged over all target classes; higher CA and lower BA are better. Small numbers show the change from \textit{Without Defense} (green: better, red: worse), and bold marks the best defense per column. DePLe gives the lowest BA in nearly all settings while largely preserving CA, whereas input transformations reduce BA only partially and often lower CA.}
\label{tab:defense_results_xray}
\vspace{0.5em}
\fontsize{7.5pt}{9pt}\selectfont 
\setlength{\tabcolsep}{1pt}
\renewcommand{\arraystretch}{0.95}
\newcolumntype{Y}{>{\centering\arraybackslash}X}
\renewcommand{\dc}[3]{\begin{tabular}[t]{@{}c@{}}#1\\[-3pt]{\fontsize{6pt}{6.5pt}\selectfont\textcolor{#2}{#3}}\end{tabular}}
\begin{tabularx}{\textwidth}{l *{6}{Y} p{6pt} *{6}{Y}}
\toprule
& \multicolumn{6}{c}{\textbf{MedCLIP}} && \multicolumn{6}{c}{\textbf{BioMedCLIP}} \\
\cmidrule(lr){2-7} \cmidrule(lr){9-14}
& \multicolumn{2}{c}{COVID} & \multicolumn{2}{c}{RSNA18} & \multicolumn{2}{c}{MIMIC}
&& \multicolumn{2}{c}{COVID} & \multicolumn{2}{c}{RSNA18} & \multicolumn{2}{c}{MIMIC} \\
\cmidrule(lr){2-3} \cmidrule(lr){4-5} \cmidrule(lr){6-7}
\cmidrule(lr){9-10} \cmidrule(lr){11-12} \cmidrule(lr){13-14}
Defense & CA & BA & CA & BA & CA & BA && CA & BA & CA & BA & CA & BA \\
\midrule
Without Defense (BAPLe)
  & 0.813 & 0.996 & 0.610 & 0.995 & 0.472 & 0.991
  && 0.841 & 1.000 & 0.636 & 0.999 & 0.368 & 0.996 \\
\midrule
Gaussian Blur
  & \dc{0.820}{good}{$\uparrow$0.007} & \dc{0.599}{good}{$\downarrow$0.397} & \dc{0.599}{bad}{$\downarrow$0.011} & \dc{0.518}{good}{$\downarrow$0.477} & \dc{0.470}{bad}{$\downarrow$0.002} & \dc{0.556}{good}{$\downarrow$0.435}
  && \dc{0.819}{bad}{$\downarrow$0.022} & \dc{0.914}{good}{$\downarrow$0.086} & \dc{0.600}{bad}{$\downarrow$0.036} & \dc{0.733}{good}{$\downarrow$0.266} & \dc{0.353}{bad}{$\downarrow$0.015} & \dc{0.821}{good}{$\downarrow$0.175} \\
JPEG-30
  & \dc{0.796}{bad}{$\downarrow$0.017} & \dc{0.551}{good}{$\downarrow$0.445} & \dc{0.590}{bad}{$\downarrow$0.020} & \dc{0.460}{good}{$\downarrow$0.535} & \dc{0.460}{bad}{$\downarrow$0.012} & \dc{0.503}{good}{$\downarrow$0.488}
  && \dc{0.738}{bad}{$\downarrow$0.103} & \dc{0.884}{good}{$\downarrow$0.116} & \dc{0.576}{bad}{$\downarrow$0.060} & \dc{0.794}{good}{$\downarrow$0.205} & \dc{0.328}{bad}{$\downarrow$0.040} & \dc{0.837}{good}{$\downarrow$0.159} \\
Feature Squeeze (MF-3)
  & \dc{0.789}{bad}{$\downarrow$0.024} & \dc{0.522}{good}{$\downarrow$0.474} & \dc{0.603}{bad}{$\downarrow$0.007} & \dc{0.657}{good}{$\downarrow$0.338} & \dc{0.462}{bad}{$\downarrow$0.010} & \dc{0.587}{good}{$\downarrow$0.404}
  && \dc{\textbf{0.849}}{good}{$\uparrow$0.008} & \dc{0.973}{good}{$\downarrow$0.027} & \dc{\textbf{0.631}}{bad}{$\downarrow$0.005} & \dc{0.964}{good}{$\downarrow$0.035} & \dc{\textbf{0.368}}{black!45}{$\pm$0.000} & \dc{0.965}{good}{$\downarrow$0.031} \\
Pixel Deflection
  & \dc{0.662}{bad}{$\downarrow$0.151} & \dc{0.638}{good}{$\downarrow$0.358} & \dc{0.343}{bad}{$\downarrow$0.267} & \dc{\textbf{0.318}}{good}{$\downarrow$0.677} & \dc{0.325}{bad}{$\downarrow$0.147} & \dc{0.476}{good}{$\downarrow$0.515}
  && \dc{0.491}{bad}{$\downarrow$0.350} & \dc{0.568}{good}{$\downarrow$0.432} & \dc{0.407}{bad}{$\downarrow$0.229} & \dc{0.514}{good}{$\downarrow$0.485} & \dc{0.225}{bad}{$\downarrow$0.143} & \dc{0.539}{good}{$\downarrow$0.457} \\
\midrule
\rowcolor{soft}
{DePLe (Ours)}
  & \dc{\textbf{0.833}}{good}{$\uparrow$0.020} & \dc{\textbf{0.051}}{good}{$\downarrow$0.945} & \dc{\textbf{0.680}}{good}{$\uparrow$0.070} & \dc{0.334}{good}{$\downarrow$0.661} & \dc{\textbf{0.505}}{good}{$\uparrow$0.033} & \dc{\textbf{0.192}}{good}{$\downarrow$0.799}
  && \dc{0.834}{bad}{$\downarrow$0.007} & \dc{\textbf{0.544}}{good}{$\downarrow$0.456} & \dc{0.577}{bad}{$\downarrow$0.059} & \dc{\textbf{0.308}}{good}{$\downarrow$0.691} & \dc{0.359}{bad}{$\downarrow$0.009} & \dc{\textbf{0.425}}{good}{$\downarrow$0.571} \\
\bottomrule
\end{tabularx}
\end{table*}

\begin{table*}[t]
\centering
\caption{Defense results on backdoored histopathology models (conventions as in Table~\ref{tab:defense_results_xray}). DePLe gives the lowest BA in most settings and improves CA throughout, whereas input transformations vary across models and datasets.}
\label{tab:defense_results_histo}
\vspace{0.5em}
\fontsize{7.5pt}{9pt}\selectfont 
\setlength{\tabcolsep}{1pt}
\renewcommand{\arraystretch}{0.95}
\newcolumntype{Y}{>{\centering\arraybackslash}X}
\renewcommand{\dc}[3]{\begin{tabular}[t]{@{}c@{}}#1\\[-3pt]{\fontsize{6pt}{6.5pt}\selectfont\textcolor{#2}{#3}}\end{tabular}}
\begin{tabularx}{\textwidth}{l *{6}{Y} p{6pt} *{6}{Y}}
\toprule
& \multicolumn{6}{c}{\textbf{PLIP}} && \multicolumn{6}{c}{\textbf{QuiltNet}} \\
\cmidrule(lr){2-7} \cmidrule(lr){9-14}
& \multicolumn{2}{c}{Kather} & \multicolumn{2}{c}{PanNuke} & \multicolumn{2}{c}{DigestPath}
&& \multicolumn{2}{c}{Kather} & \multicolumn{2}{c}{PanNuke} & \multicolumn{2}{c}{DigestPath} \\
\cmidrule(lr){2-3} \cmidrule(lr){4-5} \cmidrule(lr){6-7}
\cmidrule(lr){9-10} \cmidrule(lr){11-12} \cmidrule(lr){13-14}
Defense & CA & BA & CA & BA & CA & BA && CA & BA & CA & BA & CA & BA \\
\midrule
Without Defense (BAPLe)
  & 0.878 & 0.982 & 0.834 & 0.964 & 0.902 & 0.968
  && 0.893 & 0.955 & 0.810 & 0.905 & 0.903 & 0.820 \\
\midrule
Gaussian Blur
  & \dc{0.859}{bad}{$\downarrow$0.019} & \dc{0.662}{good}{$\downarrow$0.320} & \dc{0.825}{bad}{$\downarrow$0.009} & \dc{0.864}{good}{$\downarrow$0.100} & \dc{0.892}{bad}{$\downarrow$0.010} & \dc{0.887}{good}{$\downarrow$0.081}
  && \dc{0.882}{bad}{$\downarrow$0.011} & \dc{0.543}{good}{$\downarrow$0.412} & \dc{0.810}{black!45}{$\pm$0.000} & \dc{0.833}{good}{$\downarrow$0.072} & \dc{0.908}{good}{$\uparrow$0.005} & \dc{0.711}{good}{$\downarrow$0.109} \\
JPEG-30
  & \dc{0.846}{bad}{$\downarrow$0.032} & \dc{0.555}{good}{$\downarrow$0.427} & \dc{0.827}{bad}{$\downarrow$0.007} & \dc{0.742}{good}{$\downarrow$0.222} & \dc{0.749}{bad}{$\downarrow$0.153} & \dc{0.874}{good}{$\downarrow$0.094}
  && \dc{0.775}{bad}{$\downarrow$0.118} & \dc{\textbf{0.453}}{good}{$\downarrow$0.502} & \dc{0.723}{bad}{$\downarrow$0.087} & \dc{0.687}{good}{$\downarrow$0.218} & \dc{\textbf{0.920}}{good}{$\uparrow$0.017} & \dc{\textbf{0.602}}{good}{$\downarrow$0.218} \\
Feature Squeeze (MF-3)
  & \dc{0.857}{bad}{$\downarrow$0.021} & \dc{0.819}{good}{$\downarrow$0.163} & \dc{0.829}{bad}{$\downarrow$0.005} & \dc{0.918}{good}{$\downarrow$0.046} & \dc{0.882}{bad}{$\downarrow$0.020} & \dc{0.937}{good}{$\downarrow$0.031}
  && \dc{0.893}{black!45}{$\pm$0.000} & \dc{0.775}{good}{$\downarrow$0.180} & \dc{0.791}{bad}{$\downarrow$0.019} & \dc{0.865}{good}{$\downarrow$0.040} & \dc{0.894}{bad}{$\downarrow$0.009} & \dc{0.736}{good}{$\downarrow$0.084} \\
Pixel Deflection
  & \dc{0.259}{bad}{$\downarrow$0.619} & \dc{0.426}{good}{$\downarrow$0.556} & \dc{0.610}{bad}{$\downarrow$0.224} & \dc{0.966}{bad}{$\uparrow$0.002} & \dc{0.539}{bad}{$\downarrow$0.363} & \dc{0.998}{bad}{$\uparrow$0.030}
  && \dc{0.563}{bad}{$\downarrow$0.330} & \dc{0.465}{good}{$\downarrow$0.490} & \dc{0.738}{bad}{$\downarrow$0.072} & \dc{0.795}{good}{$\downarrow$0.110} & \dc{0.492}{bad}{$\downarrow$0.411} & \dc{0.991}{bad}{$\uparrow$0.171} \\
\midrule
\rowcolor{soft}
{DePLe (Ours)}
  & \dc{\textbf{0.943}}{good}{$\uparrow$0.065} & \dc{\textbf{0.212}}{good}{$\downarrow$0.770} & \dc{\textbf{0.837}}{good}{$\uparrow$0.003} & \dc{\textbf{0.466}}{good}{$\downarrow$0.498} & \dc{\textbf{0.907}}{good}{$\uparrow$0.005} & \dc{\textbf{0.369}}{good}{$\downarrow$0.599}
  && \dc{\textbf{0.962}}{good}{$\uparrow$0.069} & \dc{0.666}{good}{$\downarrow$0.289} & \dc{\textbf{0.869}}{good}{$\uparrow$0.059} & \dc{\textbf{0.516}}{good}{$\downarrow$0.389} & \dc{0.905}{good}{$\uparrow$0.002} & \dc{0.664}{good}{$\downarrow$0.156} \\
\bottomrule
\end{tabularx}
\end{table*}